\documentclass[11pt,a4paper]{article}
\usepackage[hyperref]{eacl2021}
\usepackage{times}
\usepackage{latexsym}

\usepackage{multirow}
\usepackage{booktabs}       
\usepackage{amsmath}

\usepackage{microtype}

\aclfinalcopy 

\title{A Comparison of Approaches to Document-level Machine Translation}

\author{Zhiyi Ma \quad Sergey Edunov \quad Michael Auli \\
  Facebook AI Research \\
  \texttt{\{mazhiyi, edunov, michaelauli\}@fb.com}
}

\date{}

\begin{document}
\maketitle
\begin{abstract}
Document-level machine translation conditions on surrounding sentences to produce coherent translations. 
There has been much recent work in this area with the introduction of custom model architectures and decoding algorithms. 
This paper presents a systematic comparison of selected approaches from the literature on two benchmarks for which document-level phenomena evaluation suites exist. 
We find that a simple method based purely on back-translating monolingual document-level data performs as well as much more elaborate alternatives, both in terms of document-level metrics as well as human evaluation.
\end{abstract}


\section{Introduction}

Machine translation made a lot of progress with the invention of better model architectures~\citep{bahdanau2015neural,gehring2017convs2s,vaswani2017transformer} and data augmentation techniques~\citep{sennrich2016bt,edunov2018bt}. 
This has been followed by reports of model performance approaching or exceeding human-level accuracy~\citep{wu2016google,hassan2018parity}. 
However, these claims usually only hold when models are evaluated on the sentence-level and have been disproven when the same translations are evaluated on the document-level~\citep{toral2018wmt}.  

Neural networks have made it easier to incorporate more context into models compared to engineered features.
For example, modern language models are able to exploit long-range context by obtaining better perplexity through conditioning predictions on entire Wikipedia articles instead of the limited context provided by individual sentences~\citep{merity2016pointer,baevski2019adaptive,dai2019xl}.

For machine translation, there has been a lot of interest in document-level translation resulting in custom architectures~\citep{jean-etal-2015-using,jean2017does,zhang2018improving}, data augmentation techniques to address the scarcity of parallel data with document boundaries~\citep{junczys2019wmt,voita2019context}, and better decoding algorithms to capture document context with language models~\citep{yu2020better}.

In this paper, we present a comparison of various approaches and evaluate their performance in a common setup in terms of BLEU, document-level consistency metrics, as well as human judgments.
Our work complements another recent comparison study~\citep{lopes2020document} by focusing on methods that leverage additional monolingual data.

Experimental results show that a simple baseline trained only on back-translated document-level data can perform very competitively compared to both DocRepair~\citep{voita2019context} and neural noisy channel modeling~\citep{yu2020better}, two recently introduced document-level approaches leveraging monolingual data in a much more elaborate and compute-intensive way.

\section{Related work}

There has been a lot of work on custom model architectures to integrate document context into translation models. 
Most work focuses on improving context representations, such as context-aware encoders~\citep{voita2018context, zhang2018improving}, context-aware decoders~\citep{voita2019good}, and hierarchical history representations~\citep{wang2017exploiting,miculicich2018document}, as well as the application of memory networks~\citep{maruf2018document}.
Pretraining has also been shown to be effective for document-level translation~\citep{liu2020multilingual}. 

Since document-level bitext is scarce, there have been several studies on applying data augmentation methods using monolingual document-level data. 
Adding monolingual data can be effective, either by creating synthetic bitext using back-translation~\citep{junczys2019wmt} or learning to correct inconsistencies on the document-level using round-trip translations~\citep{voita2019context}. 
Noisy channel modeling~\citep{yu2017neuralnoisy, yee2019simple} has also been applied to make use of language models trained on monolingual documents to capture cross-sentence context~\citep{yu2020better}.

Evaluation of document-level machine translation is also an active area of research. \citet{scherrer2019analysing} studied the effect on translation quality by manipulating discourse-level properties in data, and document-level consistency metrics on test sets~\citep{voita2019good, muller2018large} have been promoted as a good indicator on document-level translation quality.

\citet{lopes2020document} presents a systematic study on document-level translation methods which focuses on model architectures.
We complement their work by also studying methods which leverage additional monolingual document-level data.

\section{Methods}

Next, we outline a number of approaches to document-level translation as well as a few simple baselines.
\autoref{tab:comparison_methods} provides an overview and shows the amount of context modeled by each method.

\begin{table*}[t]
    \centering
    \begin{tabular}{l r r}
         \toprule 
         Method & source context & target context \\
         \midrule 
         Sent & sent & sent \\
         SentBT & sent & sent \\
         NoisyChannelSent & sent & sent \\ 
         \midrule 
         Doc2Sent & doc & sent \\ 
         Window2Window & left doc & left doc \\ 
         Doc2Doc & doc & left doc \\
         \midrule
         DocRepair & sent & doc \\
         DocBT & doc & left doc \\
         NoisyChannelDoc & sent/doc & left doc \\ 
         \bottomrule
    \end{tabular}
    \caption{Overview of approaches and the context modeled by each. We compare pure sentence-level methods, document-level techniques, both with and without data augmentation to leverage monolingual data.}
    \label{tab:comparison_methods}
\end{table*}

\subsection{Sentence-level baselines}

\paragraph{Sentence-level baseline (Sent). } 
This is a standard sequence to sequence model trained on pairs of individual sentences and this approach does not model any document-level context.

\paragraph{Back-translation (SentBT).}
To improve the sentence-level baseline (Sent) we consider back-translation as a stronger baseline.
Back-translation uses additional target-side monolingual data to improve machine translation models~\citep{bojar2011bt_pbmt,sennrich-etal-2016-improving,edunov2018bt}.
This is done by generating a synthetic source for the target monolingual data via a model trained to translate from the target language to the source language.
We denote training with just synthetic sentence-level data as SentBT, and including true bitext sentence-level data as Sent + SentBT.

\subsection{Context-aware approaches} \label{sec:contextaware}

\paragraph{Doc2Sent. } 
In this setup, the model takes as input both a document and a mask specifying which source sentence is to be translated.
It then translates this specific target sentence only. To translate the next source sentence, the mask is changed and so forth.
The mask is implemented as a learnable embedding representing binary values for each token and it is added to the encoder output.
Doc2Sent models the full source document, but no context beyond the current target sentence prefix is modeled on the target side.
This setup enables us to understand the importance of having target-side document context.


\paragraph{Window2Window. } 
This models a sliding window of a fixed number of sentences $N$, both in the source and the target.
The window is adjusted after a source sentence is translated and we concatenate the last generated sentence to form the final generation, i.e., the previous $N-1$ sentences are treated as context only. In our experiments we set $N=2$. 

\paragraph{Doc2Doc. }
\citet{tiedemann2017context} and then~\citet{junczys2019wmt} proposed a simple but very effective document-level translation approach by training a standard sequence to sequence model on pairs of bitext documents.
Documents are split into examples of no more than 1,000 subword units and sentences are separated with a special token. 
This makes the sometimes incorrect assumption that the number of sentences in parallel documents is the same.
There is no alignment of sentences when the document is split into smaller chunks but~\citet{junczys2019wmt} finds that often the correct number of target sentences is predicted at inference time.
The method outperforms very strong sentence-level systems in human evaluation.
Doc2Doc models the full source context as well as the target document prefix.

\paragraph{Back-translation (DocBT). } 
We can also back-translate monolingual document-level data in the target language with a sentence-level system~\citep{junczys2019wmt}.
The resulting source sentences are simply concatenated into a source document and the source and target document pair is treated as a training example.
We denote training with just synthetic data as DocBT, and including true bitext document-level data as (Doc2Doc + DocBT).

\subsection{DocRepair}
\citet{voita2019context} proposed a post-editing method to fix inconsistencies between sentence-level translations using target side document-level monolingual data. 
This method requires sentence-level translation models for both forward and backward directions.
At training time, monolingual target documents are translated to the source and then back to the target using the sentence-level systems.
This yields groups of round-trip translated sentences from the original monolingual document-level data which may contain inconsistencies with respect to each other.

Another model is trained to map these inconsistent groups of sentences to the original consistent documents (DocRepair model).
At test time, source documents are first translated into inconsistent target sentences using a sentence-level model.
This is followed by the DocRepair model mapping the inconsistent target sentence translations to a coherent document.
Since the first step is performed by a sentence-level system, this method does not use any source document context, however, DocRepair has access to the full context on the target side.

\subsection{Noisy channel modeling}
\paragraph{NoisyChannelSent.} 
By leveraging Bayes' rule, this approach models a mapping from the source $x$ to target $y$, i.e., 
$$
    p(y\vert x) = p(x\vert y)p(y)/p(x) \propto p(x\vert y)p(y)
$$
where $p(x|y)$ and $p(y)$ are referred to as the channel model and language model, respectively. 
Standard sequence to sequence models directly parameterize $p(y|x)$, which is referred to as direct model.
As another baseline, we train the channel model and the language model on sentence-level data~\citep{yu2017neuralnoisy,yee2019simple} and rerank the n-best list output of a sentence-level direct model.
For reranking, we choose the hypothesis which maximizes  $\frac{1}{t}\log(y\vert x) + \frac{\lambda_1}{s}\left(\log p(x\vert y) + \log p(y)\right)$ where $t$ is the length of target, $s$ is the source length and $\lambda$ is a tunable weight~\citep{yee2019simple}.


\paragraph{NoisyChannelDoc. } 
\citet{yu2020better} proposed a context-aware noisy channel reranking method by using document-level language models. 
First, n-best lists for every source sentence are generated, either with a sentence-level direct model or a document-level direct model. 
These n-best lists are then reranked with a beam-search that applies the document-level language model as well as a sentence-level channel model. 
The beam search maximizes
\begin{equation}
  \begin{aligned}
    &\lambda_1\log q(\textbf{y}^{\leq i}\vert \textbf{x}^{\leq i}) + \lambda_2\log p(\textbf{x}^{\leq i}\vert \textbf{y}^{\leq i}) \\ 
    &+ \log p(\textbf{y}^{\leq i}) + \lambda_3 \vert \textbf{y}^{\leq i}\vert
  \end{aligned}
\end{equation}
where $\textbf{x}^{\leq i}$ and $\textbf{y}^{\leq i}$ are partial source and target documents and $\vert . \vert$ denotes the total number of tokens; $\lambda_1, \lambda_2, \lambda_3$ are hyper-parameters to be optimized. 
Depending on the direct model which performed the initial translation, this approach uses the entire source context or the current source sentence. The target document-level language model uses the entire target prefix.

\section{Experimental setup}
\label{sec:setup}

We perform experiments on two benchmarks, OpenSubtitles2018 English-Russian (en-ru) and WMT'17 English-German (en-de) and next we outline these benchmarks, the evaluation protocol as well as the model setup.

\subsection{Datasets}
\label{sec:datasets}

\paragraph{Opensubtitles2018 English-Russian (en-ru).} 
For this dataset we follow the setup of~\citet{voita2018context} and~\citet{voita2019context}. 
The corpus has been originally derived from publicly available OpenSubtitles2018~\citep{lison2019open} English and Russian data, and consists of three parts: 6M sentence-level bitext examples, 1.5M bitext documents and 30M monolingual Russian documents.
The document-level parallel corpus comprises 1.5M examples of four sentences each (denoted as 1.5M$_d$).
Examples are based on a sliding window over 2M unique sentences, where sentences are a subset of the 6M sentence-level bitext. 
The monolingual data consists of 30M documents, each consisting of four consecutive Russian sentences (denoted as 30M$_{md}$), when split into sentences this monolingual data comprises 120M examples  (denoted as 120M$_m$).  

The validation and test sets contain 10k documents of four sentences each, constructed similarly to the training data but heldout to avoid overlap. There is no overlap between training bitext or monolingual data with validation and test set on the document-level.

We use byte pair encoding (BPE; \citealt{sennrich2016neural}) with 24k merges to segment words into subword units for English and Russian, respectively. The dataset is already tokenized and we compute tokenized case-insensitive BLEU on the document-level.\footnote{Dataset available at \url{https://github.com/lena-voita/good-translation-wrong-in-context}}
BLEU is computed with Moses \texttt{multi-bleu.perl}~\citep{koehn2007moses}. 


\paragraph{WMT17 English-German (en-de).} 
For this benchmark, we follow the setup of~\citet{muller2018large} whose training data includes the Europarl, Common Crawl, News Commentary and Rapid corpora, totaling nearly 6M sentence pairs.

As monolingual data we use Newscrawl 2017 in German which has document boundaries.
It contains 3.6M documents or 73M sentences (denoted as 73M$_m$). 
We preprocess the data by normalizing punctuation, removing non-printable characters and Moses tokenization~\citep{koehn2007moses}.
We use BPE with 32k merges shared between English and German. We used shared vocabularies for English and German because these performed best in initial experiments. 
To be consistent with the English-Russian document-level setup, we split each document of the monolingual data into separate examples of up to four sentences or a maximum of 1000 tokens. 
This results in a total of 19.5M documents (denoted as 19.5M$_{md}$).

The bitext training data does not provide explicit document boundaries but most of the sentences are ordered as in the original documents. 
We consider a setup, where we simply split the bitext training data into pseudo-documents following the same strategy as for the monolingual data.
This results in 1.3M documents of up to four sentences or 1k tokens each (denoted as 1.3M$_d$).

We use newstest2016 for validation, containing 155 documents with 2999 sentences. 
As test set we use newstest2017, newstest2018 and newstest2019,
which contain 130 documents with 3004 sentences, 122 documents with 2998 sentences and 123 documents with 1997 sentences, respectively.
We split the data into separate examples, similar to the monolingual data, leading to 811 documents for the validation data, and 796, 799 and 549 documents for each test set, respectively.
We evaluate with detokenized BLEU~\citep{post2018call}

\subsection{Contrastive evaluation}

In order to capture translation quality on the document-level, we consider two consistency evaluation sets which are available for the two benchmarks we consider.
Both evaluation suites require distinguishing sentences which are consistent with the provided context, typically several preceeding sentences.
This is done by simply choosing the sentence which obtains the highest model score according to the translation model by scoring the possible translations provided by the challenge set.

\paragraph{Discourse phenomena for English-Russian.} 
\citet{voita2019good} evaluate four discourse phenomena for English-Russian, namely deixis, lexical cohesion, ellipsis infl. and ellipsis verb prediction (VP). 
We provide a short overview of this test set and refer the reader to the original paper for further details.
The deixis evaluation requires discriminating between formal and informal Russian translations of the English \textit{you} depending on the context.
Lexical cohesion focuses on consistent translation of named entities across the entire document.
Ellipsis evaluates whether the translation model can disambiguate elliptical structures.
Ellipsis inflection (infl.) evaluates whether models can correctly predict the morphological form of a noun group which can only be understood from context beyond that sentence. 
Ellipsis VP tests for the correct translation of a verb phrase that does not exist in Russian. 
The total number of examples in each test set is 3000, 2000, 500, 500 for deixis, lexical cohesion, ellipsis infl. and ellipsis VP, respectively. 
All examples are four sentences long.

\paragraph{Pronoun translation for English-German.} 
\citet{muller2018large} presents a large scale contrastive test set for pronoun translation in English-German which requires document-level context.
It tests the ability to identify the correct German translation of the English pronoun \textit{it} as either \textit{es}, \textit{sie} and \textit{er}.
The evaluation set contains 12k examples with 4k for each pronoun. The number of context sentences is customizeable, and for 80\% of test examples, document-level context is required to produce the correct translation.
For sentence-level models, we use no context, and for document-level models, we use the number of available sentences in our documents, which is typically four for WMT17 en-de.

\subsection{Human Evaluation}
\label{sec:setup_human}


Human evaluations were performed by certified professional translators that are native speakers of the target language as well as fluent in the source language.
All assessments are conducted on the document-level, using exactly the same data as used for document-level models, as described in Section~\ref{sec:datasets}.
To compare multiple systems in English-Russian we use source based direct assessment. 
Raters evaluate correctness and completeness on a scale of 1-100 for each translation given a source document. 
This evaluation has the benefit of being independent of the provided human references which may affect the evaluation. 

We collected three judgements per translation. If any two raters disagree by more than 30 points, we discard the result and request reevaluation of the translation. 
Evaluation was blind and randomized: human raters did not know the identity of the systems and all outputs were shuffled to ensure that each rater provides a similar number of judgements for each system.

Following the WMT shared task evaluation protocol~\citep{bojar2018wmt}, we normalize the scores of each rater by the mean and standard deviation of all ratings provided by the rater. We remove raters who have rated fewer than 10 translations in total. 
Next, we average the normalized ratings for each sentence and average all per-translation scores to produce an aggregate per-system z-score.
We randomly sampled 200 examples from the standard test set and 100 examples from the consistency test set (25 from each discourse phenomenon), and conducted human evaluation for the two sets independently.

We confirm our findings on English-German, for which we did a system comparison study to directly compare a few select systems. 
Human annotators were presented with a source document and two candidate translations and were asked to judge which translation is better. 
For each translation, we collect three judgements and determine human preference based on the system which is preferred by the majority of raters.

\subsection{Model training}

Models are implemented in \texttt{fairseq}~\citep{ott2019fairseq}. 
We use the Adam optimizer~\citep{kingma2015adam}, with $\beta_1 = 0.9$, $\beta_2 = 0.98$ and $\epsilon = 10^{-9}$. 
We use the learning rate schedule described in~\citet{vaswani2017transformer} with 4,000 warmup steps, an initial learning rate of $10^{-7}$ and a maximum learning rate of $5\times 10^{-4}$.

\paragraph{OpenSubtitles2018 English-Russian.}
We use transformer base models with dropout 0.3, train for 300k updates on 8 GPUs and tune the batch size on the validation set in the range of 128k and 512k tokens. 
We use early stopping when validation loss stops improving and apply checkpoint averaging on last 5 checkpoints. 
For generation, we use beam search of width 4, following~\citep{voita2019context}, and tune the length penalty on the validation data.\footnote{We tried the following values: [0.01, 0.1, 0.3, 0.5, 0.8, 1, 2, 4, 8]}.

\paragraph{WMT17 English-German.} 
We train transformer big models for 300k updates on 32 GPUs with a batch size of 262k tokens, and early stop based on the validation loss. 
We use the checkpoint with the best validation loss without averaging.
For generation we use a beam width of 5 and tune the length penalty on the same set of values as English-Russian.

\paragraph{Language model.} 
We use a transformer big decoder only~\citep{baevski2019adaptive}, with 12 decoder layers, dropout 0.1, embedding dimension 512, and without layer normalization~\citep{ba2016layer} after the last decoder block. 
We use a cosine learning rate scheduler where the learning rate is increased linearly from $10^{-7}$ to $1$ for 16k warmup steps~\citep{loshchilov2016sgdr}.
We tune the number of updates in the range [316k, 616k, 916k], use the best checkpoint according to validation loss, and train on 8 GPUs with a batch size of 16k tokens for English-Russian and on 32 GPUs with a batch size of 65.5k tokens for English-German.

\paragraph{Back-translation.} 
Synthetic sources are generated with an ensemble of four models and unrestricted sampling~\citep{edunov2018bt}.
For models trained with a combination of true bitext and back-translated data, we upsample the true bitext by tuning the upsample ratio over the values [1, 10, 20, 40, 60].

\section{Results}

\begin{table*}[t]
    \small
    \centering
    \begin{tabular}{ll|c|cccccc}
        \toprule 
        Method & Training data & BLEU ($\uparrow$) & \multicolumn{5}{c}{Consistency test set ($\uparrow$)} \\
        & & & avg & deixis & lex. c. & ell. infl. & ell. VP \\
        \midrule 
        Sent~\citep{voita2019context} & 6M & 33.9 & 44.3 & 50.0 & 45.9 & 53.0 & 28.4 \\
        DocRepair~\citep{voita2019context} & 6M + 30M$_{md}$ & 34.6 & 83.5 & 91.8 & 80.6 & 86.4 & 75.2\\
        \midrule 
        Sent & 6M & 34.1 & 44.5 & 50.0 & 45.9 & 54.6 & 27.6 \\
        + SentBT & + 120M$_{m}$ & 35.2 & 43.1 & 40.5 & 47.2 & 39.4 & 45.2 \\
        NoisyChannelSent & 6M + 120M$_{m}$ & 35.5 & 44.3 & 50.0 & 45.9 & 55.2 & 26.2 \\
        \midrule
        Doc2Sent & \multirow{3}{*}{1.5M$_{d}$} & 33.0 & 55.4 & 50.1 & 45.9 & 56.2 & 69.2 \\
        Window2Window & & 33.2 & 65.3 & 63.5 & 46.1 & 80.4 & 71.2 \\
        Doc2Doc & & 32.7 & 77.3 & 88.6 & 54.9 & 82.6 & 83.0 \\
        \midrule
        DocRepair & 6M + 30M$_{md}$ & 34.6 & 85.8 & 92.1 & \textbf{81.2} & \textbf{90.2} & 79.6 \\
        NoisyChannelDoc & 6M + 30M$_{md}$ & \textbf{36.0} & 58.2 & 66.6 & 67.1 & 57.6 & 41.4 \\ 
        + Doc2Sent & + 1.5M$_d$ & 35.2 & 66.9 & 65.1 & 70.8 & 60.4 & 71.2 \\ 
        DocBT & 6M + 30M$_{md}$ & 34.8 & \textbf{87.1} & \textbf{93.6} & 79.6 & 89.8 & 85.2 \\
        + Doc2Doc & + 1.5M$_{d}$ & 35.0 & 87.0 & 92.5 & 80.3 & 89.2 & \textbf{86.0} \\
        \bottomrule
    \end{tabular}
    \caption{Results on Opensubtitles 2018 English to Russian translation in terms of BLEU on the test set and consistency evaluation scores~\citep{voita2019context}. (Please note the avg column is an arithmetic average of the four discourse phenomena.)
    We indicate the amount and type of training data: no subscript denotes sentence-level bitext, (m) denotes monolingual data, (d) denotes document-level data.
    }
    \label{tab:en_ru_bleu}
\end{table*}

\begin{table*}[t]
    \centering
    \begin{tabular}{l|ccc|ccc}
        \toprule 
        Method & \multicolumn{3}{c}{Test set} & \multicolumn{3}{c}{Consistency test set} \\
        & Z scores ($\uparrow$) & std & rank & Z scores ($\uparrow$) & std & rank \\
         \midrule 
        Reference & \textbf{0.234*} & 0.057 & 1 &  \textbf{0.378*} & 0.061 & 1  \\
        \midrule
        Sent & 0.055 & 0.051 & 6  & -0.023 & 0.086 & 7  \\
        + SentBT & 0.113 & 0.053 & 2  & -0.074 & 0.081 & 8  \\
        NoisyChannelSent & 0.088 & 0.058 & 4 & -0.017 & 0.092 & 6 \\
        \midrule
        DocRepair & 0.026 & 0.064 & 7  & 0.118 & 0.1 & 5 \\
        DocBT & 0.085 & 0.048 & 5  & \textbf{0.268*} & 0.08 & 2  \\
        Doc2Doc + DocBT & 0.014 & 0.062 & 8 & 0.195 & 0.093 & 3  \\
        NoisyChannelDoc & 0.111 & 0.046 & 3 & 0.144 & 0.093 & 4 \\ 
        \bottomrule
    \end{tabular}
    \caption{Human evaluation results for Opensubtitles 2018 English-Russian translation on the test set of the benchmark as well as on the consistency test set of~\citet{voita2019context}. We randomly sampled 200 examples from the standard test set, and 100 examples from the consistency test set (25 from each discourse phenomenon subset). Results marked with \textbf{*} are statistically significantly better than the baseline (Sent) system at p=0.05. 
    }
    \label{tab:en_ru_human_eval}
\end{table*}

\subsection{English-Russian translation}

We first compare various approaches to document-level translation as well as sentence-level baselines on the English-Russian Opensubtitles 2018 benchmark (\autoref{tab:en_ru_bleu}).
We measure BLEU on the document-level test set of Opensubtitles and accuracy on the consistency evaluation.

First, we find that sentence-level systems perform well in terms of BLEU but poorly in terms of the document-level consistency evaluation. 
This includes a system trained purely on sentence-level data (Sent), augmented with back-translated data (Sent + SentBT), and noisy channel reranking with a sentence-level language model (NoisyChannelSent; \citealt{yu2017neuralnoisy,yee2019simple}).

Second, we evaluate document-level systems trained purely on bilingual document-level training data (1.5M documents) to understand how the amount of context modeled impacts accuracy.
In terms of the consistency evaluation, we find that these systems perform better with more context modeled: 
Doc2Sent uses the entire source context but models only a single target sentence. 
This performs least well, although better than the sentence-level baselines. 
Modeling a sliding window of source and target sentences improves on this (Window2Window) and
treating the entire document as a consecutive sequence performs best (Doc2Doc).

However, in terms of BLEU, all aforementioned document-level systems underperform the sentence-level systems. 
We suspect that this is because these document-level systems are trained on less bitext data: the 1.5M documents contain only 2M unique sentences since documents were created through a sliding window over the 6M bitext sentences used by Sent (\autoref{sec:datasets}).

Third, we evaluate various document-level approaches based on adding 30M monolingual documents (30M$_{md}$). 
In terms of the consistency evaluation, our reimplementation of DocRepair~\citep{voita2019context} performs very well and outperforms the sentence-level systems, including the ones based on sentence-level back-translation (SentBT).
Noisy channel reranking with a document-level language model~\citep{yu2020better} peforms very well in terms of BLEU but less so in terms of the consistency evaluation. 
The n-best lists to be reranked by the noisy channel approach are based on a sentence-level-system and we therefore re-generate them with a document-level system (Doc2Sent). 
This improves the consistency evaluation but still does not perform as well as the other approaches relying on document-level monolingual data.

Finally, simply back-translating the monolingual documents and training a standard sequence-to-sequence model on this data outperforms all above approaches on the consistency test set, including DocRepair which requires two translation steps at inference time compared to a single back-translation step for DocBT.
Interestingly, adding the true bitext documents (1.5M$_d$, Doc2Doc) does not improve over solely back-translated documents (DocBT).

Automatic evaluation in terms of BLEU and the consistency test set results are not in strong agreement. 
We therefore collect judgments from professional human translators with source-based direct assessment (\autoref{sec:setup_human}). 
For this evaluation we retain all systems except for Doc2Sent and Window2Window to make the human study more manageable and because these systems were clearly outperformed by Doc2Doc.

Human judgements (Table~\ref{tab:en_ru_human_eval}) on the documents of the consistency test set confirm that DocBT performs very well compared to the other data augmentation-based approaches (DocRepair, NoisyChannelDoc) and the results show a clear distinction between sentence-level and document-level approaches.
However, human preferences are much less pronounced on the standard test set with no systems clearly outperforming the others.
This is likely because the examples in the consistency test set were selected to test for phenomena which are not as prevalent in existing test sets.

\subsection{English-German translation} 

\begin{table*}[t]
\small
    \centering
    \begin{tabular}{ll|ccc|cccc}
        \toprule 
        Method & Training data & \multicolumn{3}{c}{BLEU ($\uparrow$)} & \multicolumn{4}{c}{Contrastive reference pronoun} \\
        & & 2017 & 2018 & 2019 & total($\uparrow$) & \textit{es} & \textit{er} & \textit{sie} \\
        \midrule 
        \citet{muller2018large} Sent & 6M & 24.6 & 35.4 & - & 0.47 & 0.81 & 0.22 & 0.38 \\
        \citet{muller2018large} Doc & 6M$_d$ & 25.3 & 36.5 & - & 0.49 & 0.84 & 0.23 & 0.39 \\
        \midrule 
        Sent & 6M & 28.3 & 41.1 & 37.9 & 0.50 & 0.86 & 0.26 & 0.39 \\
        + SentBT & + 73M$_{m}$ & 31.4 & 43.8 & 36.7 & 0.52 & 0.87 & 0.31 & 0.39\\
        NoisyChannelSent & 6M + 73M$_{m}$ & 29.7 & 43.0 & 38.2 & 0.52 & 0.86 & 0.29 & 0.41 \\
        \midrule
        Doc2Doc & 1.3M$_{d}$ & 28.7 & 41.5 & 37.9 & 0.81 & 0.92 & 0.76 & 0.76 \\
        \midrule 
        DocRepair & 6M + 19.5M$_{md}$ & 28.8 & 38.9 & 32.0 & 0.80 & 0.90 & 0.70 & 0.81 \\
        NoisyChannelDoc & 6M + 19.5M$_{d}$ & 29.9 & 43.1 & 38.4 & 0.62 & 0.90 & 0.42 & 0.54 \\ 
        DocBT & 6M + 19.5M$_{md}$ & 30.6 & 41.4 & 32.1 & 0.81 & 0.91 & 0.73 & 0.77\\
        + Doc2Doc & + 1.3M$_{d}$ & 32.8 & 45.8 & 37.7 & 0.81 & 0.92 & 0.71 & 0.79 \\
        \bottomrule
    \end{tabular}
    \caption{Results on WMT17 English to German translation in terms of BLEU on various WMT test sets, and a contrastive test suite evaluating pronoun selection~(\citealt{muller2018large}; cf.~\autoref{tab:en_ru_bleu}).
    }
    \label{tab:en_de_bleu_contrapro}
\end{table*}

\begin{table*}[t]
    \centering
    \begin{tabular}{ll|cc|c}
         \toprule 
         Test No. & Method & \multicolumn{2}{c}{Objective metric} & Human  \\
         & & BLEU ($\uparrow$) & Contrastive score total ($\uparrow$) & preference ($\uparrow$) \\
         \midrule 
         \multirow{2}{*}{1} & Sent & 37.9 & 0.50 & 0.33 \\
         & DocBT & 32.1 & 0.81 & \textbf{0.62*} \\
         \midrule
         \multirow{2}{*}{2} & DocRepair & 32.0 & 0.80 & 0.47 \\
         & DocBT & 32.1 & 0.81 & 0.48 \\
         \midrule 
         \multirow{2}{*}{3} & NoisyChannelDoc & 38.4 & 0.62 & 0.32 \\
         & DocBT & 32.1 & 0.81 & \textbf{0.60*} \\
         \bottomrule 
    \end{tabular}
    \caption{Human preferences on WMT17 English-German data. We ask human raters to indicate which system is preferred on 100 randomly sampled examples from newstest2019, each up to 1000 tokens long (\autoref{sec:datasets}). Results marked by * are statistically significantly better than the other system at $p = 0.05$.}
    \label{tab:en_de_human_eval}
\end{table*}

So far we saw that simple back-translation of documents (DocBT) performed competitively to more complicated semi-supervised methods.
To confirm these findings we perform another experiment on WMT17 English-German translation and compare DocBT to DocRepair and NoisyChannelDoc, as well as a few simpler alternatives.
Following~\citet{muller2018large}, we measure performance in terms of sentence-level detokenized BLEU on newstest2017-2019. 
We also compare to the best sentence-level and document-level results of~\citet{muller2018large} whose pronoun contrastive task we use in our study.


The results (\autoref{tab:en_de_bleu_contrapro}) show that sentence-level systems perform poorly on the document-level metrics which require modeling context information.
The document-level systems outperform the sentence-level baselines on the contrastive pronoun task and the simple DocBT method ranks amongst the best systems in the consistency evaluation. 
However, additional monolingual data does not improve the consistency evaluation over just training on bitext document data (Doc2Doc). 
NoisyChannelDoc performs less well than the other document-level methods.
This is likely because the n-best lists for reranking were generated with sentence-level direct models and using a document-level direct model would improve results (similar to NoisyChannelDoc + Doc2Sent in~\autoref{tab:en_ru_bleu}). 

Similar to before, BLEU does not enable strong conclusions. 
In particular, DocBT performs poorly on newstest2019 which is a test set that is purely forward translated, that is, sentences originally written in English are paired with German human translations and thus BLEU is measured against human translated text~\citep{bojar2019wmt}. 
This is also the case for DocRepair, whose training data involves roundtrip translation.
While realistic, for this setup, BLEU has been shown to correlate very poorly with human judgements on forward translated test data~\citep{edunov2020bt}.
We therefore also evaluate BLEU on the German-English version of newstest2019 with source and target reversed and find that DocBT and DocBT + DocDoc obtains the highest BLEU amongst all systems on this test set, followed by DocRepair.

To draw stronger conclusions about the performance of DocBT, we perform another smaller human study. 
We ask professional human translators to give preference ratings for DocBT vs. the sentence-level baseline (Sent) in a first evaluation and DocBT vs. NoisyChannelDoc in a second evaluation. 
We focus on NoisyChannelDoc in favor of DocRepair because the former achieved better BLEU.\footnote{Human evaluation for DocRepair vs DocBT is in progress and will be included in the next version of this paper.}
\autoref{tab:en_de_human_eval} clearly shows that DocBT is clearly both over the sentence-level baseline (Sent) as well as the more complicated NoisyChannelDoc method.

\section{Conclusion}


We compared several recent approaches to document-level translation on two benchmark datasets.
We find that training a standard sequence to sequence model on back-translated document-level monolingual data presents a very competitive baseline.
We encourage future research in document-level translation to compare to this baseline.

Evaluation of document-level translation is challenging and we present results in terms of automatic metrics as well as human evaluation.
Document-level consistency evaluation suites are useful and clearly distinguish systems capable of modeling long-range context from sentence-level systems. 
However, their construction likely overemphasizes phenomena which are not as frequent in other datasets.



\bibliography{anthology,master}
\bibliographystyle{acl_natbib}

\end{document}